# Research on Optimization of Natural Language Processing Model Based on Multimodal Deep Learning

Dan Sun[1], Yaxin Liang[2], Yining Yang[3], Yuhan Ma[4], Qishi Zhan[5], Erdi Gao[6]

[1]Washington University in St. Louis, USA, sun.dan@wustl.edu

[2]University of Southern California, USA, yaseen.liang@outlook.com

[3]Carnegie Mellon University, USA, yiningya@alumni.cmu.edu

[4]Johns Hopkins University, USA, yma62@jh.edu

[5]Marquette University, USA, qishizhan7@gmail.com

[6]New York University, USA, ge2093@nyu.edu

*Abstract*—This project intends to study the image representation based on attention mechanism and multimodal data. By adding multiple pattern layers to the attribute model, the semantic and hidden layers of image content are integrated. The word vector is quantified by Word2Vec method, and then evaluated by word embedding convolutional neural network. The published experimental results of the two groups were tested. The experimental results show that this method can convert discrete features into continuous characters, thus reducing the complexity of feature preprocessing. Word2Vec and natural language processing technology are integrated to achieve the goal of direct evaluation of missing image features. The robustness of image feature evaluation model is improved by using the excellent feature analysis characteristics of convolutional neural network. This project intends to improve the existing image feature identification methods and eliminate the subjective influence in the evaluation process. The findings from simulation indicate that the novel approach have developed is viable, effectively augmenting the features within the produced representations.

## I. Introduction

Significant research has been devoted to semantic-based scene description algorithms. Some studies have introduced a templating approach, which initially extracts information such as the location and category of a subject from an image, followed by populating a pre-arranged statement library with this information. The effectiveness of this approach is closely linked to the resultant understanding of the image; however, the descriptions generated tend to be simplistic and structurally uniform, often differing considerably from human-annotated sentences [1]. A challenge arises when the image and its associated statements are absent from the search database or markedly vary from the content of the image to be described, leading to descriptions that stray from the image's actual content[2]. Addressing these discrepancies has become a focal point in contemporary artificial intelligence research.

Despite the ability to concentrate on different parts of the image at various instances, accurately predicting certain non-visual words, such as articles and prepositions, remains reliant on linguistic patterns [3][4]. This limitation underscores the need for a strategy that not only enhances the speed of generation but also the semantic coherence of the sequenced words. This paper introduces the concept of a weighted order of magnitude. By training on this scale, the algorithm learns to discern when both the image and the associated word vector warrant attention and selection. In this project, a multi-modal hierarchical Long Short-Term Memory (LSTM) network is proposed, combined with an attention mechanism, to improve performance in image feature extraction[5].

## II. System algorithm design

First of all, LSTM will accept word vectors in the word vector space in batches, convert them into high-dimensional static vectors through the feature extraction of CNN, then weight the vector, and then transform it into multi-modal, and pass the hidden layer information of its previous step to the multi-modal layer [6]. The results of multi-modal operation are transformed into lexical vector space, and the probability distribution and error loss of terms are estimated by SoftMax[7].

Following the attribute layer, the multi-modal state layer is constructed, and the resulting hidden state and the information generated by the attention-adaptive attention mechanism are sent to the multi-modal layer for fusion[8]. Finally, the obtained multi-modal state data is submitted to the software layer, so as to obtain the probability distribution of the next sentence [9]. Figure 1 shows the overall network architecture.

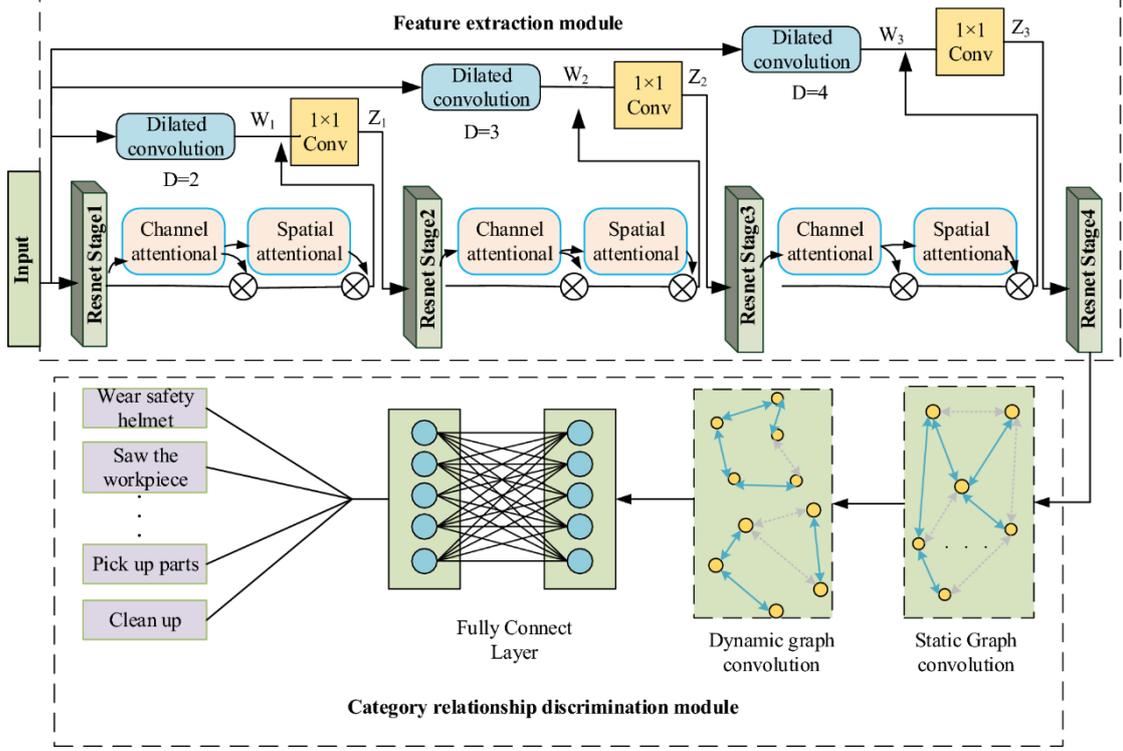

Fig. 1. Schematic diagram of the integration of attention mechanisms into neural networks

## III. WORD VECTORIZATION BASED ON WORD2VEC

The projection layer accepts all word vectors and adds them cumulatively:

$$U_\eta = \sum_{i=1}^{2z} W(Context(\eta)_i) \quad (1)$$

When dividing nodes, the category on the left side of the binary tree is negative, and the category on the right side is positive. Then according to the sigmoid function, the possibility of the node being assigned to the positive category is as follows:

$$\delta(U_\eta^T) = \frac{1}{1+e^{-U_\eta^T}} \quad (2)$$

The output layer is based on Huffman binary tree, and combines the probability product of $h^\eta - 1$ branches in path $f^\eta$ with the formula (1). The conditional probability formula and the log-likelihood degree of the conditional probability are constructed as follows:

$$f(\eta \mid Context(\eta)) = \prod_{j=2}^{h^\eta} f(s_j^\eta \mid U_\eta, \beta_{j-1}^\eta) \quad (3)$$

$$\varphi = \sum_{\eta \in \xi} f(\eta \mid Context(\eta)) \quad (4)$$

$f(s_j^\eta \mid U_\eta, \beta_{j-1}^\eta)$ is the probability of each branch of the binary tree. In combination with formula (2), the positive class probability is obtained in parallel with the negative class probability

$$f(s_j^\eta \mid U_\eta, \beta_{j-1}^\eta) = [\delta(U_\eta^T \beta_{j-1}^\eta)]^{1-s_j^\eta} \\ [1-\delta(U_\eta^T \beta_{j-1}^\eta)]^{s_j^\eta} \quad (5)$$

The log-likelihood function identity of conditional probability obtained by combining (3) ~ (5) is

$$\varphi = \sum_{\eta \in \varphi} \log \prod_{j=2}^{h_\eta} \{[\delta(U_\eta^T \beta_{j-1}^\eta)]^{1-s_j^\eta} \\ \times [1-\delta(U_\eta^T \beta_{j-1}^\eta)]^{s_j^\eta}\} = \\ \sum_{\eta \in \varphi} \sum_{j=2}^{h_\eta} \{(1-s_j^\eta)\log[\delta(U_\eta^T \beta_{j-1}^\eta)] \\ + s_j^\eta \log[1-\delta(U_\eta^T \beta_{j-1}^\eta)]\} \quad (6)$$

From formula (6), it can be seen that the log-likelihood function of conditional probability is proportional to the function in large parentheses. Let $\varphi(\eta, j)$ be the function in curly braces, so only need to optimize $\varphi(\eta, j)$, then the optimal solution of the log-likelihood function can be obtained:

$$\varphi(\eta, j) = (1-s_j^\eta)\log[\delta(U_\eta^T \beta_{j-1}^\eta)] \\ + s_j^\eta \log[1-\delta(U_\eta^T \beta_{j-1}^\eta)] \quad (7)$$

According to the stochastic gradient ascending algorithm, the optimization function requires the directional gradient of the solution function on its parameters. $\varphi(\eta, j)$ has two parameters $\beta_{j-1}^\eta$ and $U_\eta$, which are solved successively as follows:

$$\frac{\partial \varphi(\eta, j)}{\partial \beta_{j-1}^\eta} = \frac{\partial}{\partial \beta_{j-1}^\eta}\{(1-s_j^\eta)\log[\delta(U_\eta^T \beta_{j-1}^\eta)] \\ + s_j^\eta \log[1-\delta(U_\eta^T \beta_{j-1}^\eta)]\} = (1-s_j^\eta) \times \\ [1-\delta(U_\eta^T \beta_{j-1}^\eta)]U_\eta - s_j^\eta \delta(U_\eta^T \beta_{j-1}^\eta)U_\eta \\ = [1-s_j^\eta - \delta(U_\eta^T \beta_{j-1}^\eta)]U_\eta \quad (8)$$

Since $\beta_{j-1}^\eta, U_\eta$ is symmetric on a function of $\varphi(\eta, j)$, the gradient of $\dfrac{\partial \varphi(\eta, j)}{\partial U_\eta}$ is obtained from equation (8)

$$\frac{\partial \varphi(\eta, j)}{\partial U_\eta} = [1-s_j^\eta - \delta(U_\eta^T \beta_{j-1}^\eta)]\beta_{j-1}^\eta \quad (9)$$

After the output layer obtains the accumulation vector $U_\eta$ of the mapping layer, the word vector $W(\eta)$ of word $\eta$ is updated based on the relation between equations (1), (8) and (9). That is, every time parameter $U_\eta$ is updated, $W(\eta)$ is also updated, so the output of the output layer is

$$W(\eta) := W(\eta) + \kappa \sum_{j=2}^{h_\eta} \frac{\partial \varphi(\eta, j)}{\partial U_\eta} \quad (10)$$

$\kappa$ is the set learning rate; $\sum_{j=2}^{h_\eta} \dfrac{\partial \varphi(\eta, j)}{\partial U_\eta}$ indicates that after the gradient update, the cumulative sum of the mapping layer is spread back to each word vector.

IV. NEURAL NETWORK TRAINING MODEL BASED ON CNN

Deep learning is a data-based representation learning, which is superior to other machine learning algorithms in feature extraction, especially in object detection, natural language processing and so on [10][11][12]. In this paper, CNN is introduced into the credit evaluation, one-dimensional CNN is constructed under the framework of WV-CNN, and the text of lexical vectorized credit characteristics is used as input to train and evaluate it. As can be seen from Figure 2, the word vector information obtained by the CBOW algorithm to the input level of the network, that is, the word hidden layer[13]. In addition, to avoid over-fitting, Dropout layers are added between the convolution layer and the pool layer. The Flatten layer is flat after several convolution and pooling operations. Finally, the results of credit evaluation are output through the complete connectivity hierarchy.

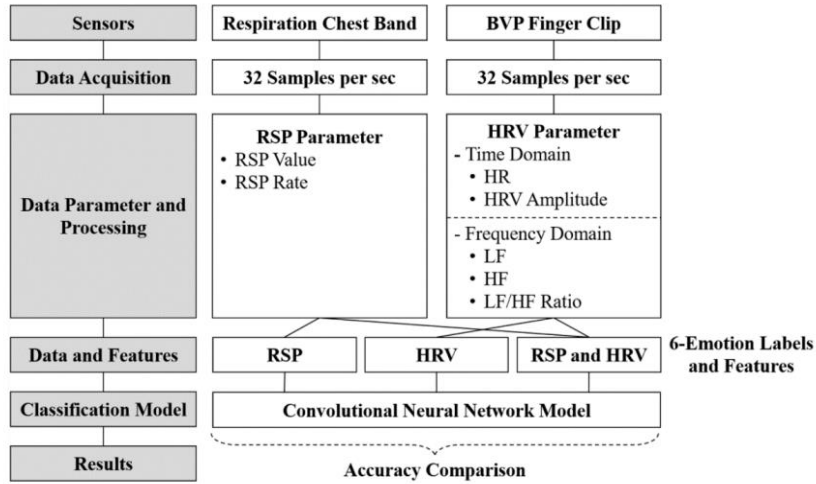

Fig. 2. CNN structure diagram in WV-CNN model

V. MODEL IMPLEMENTATION BASED ON KERAS DEEP LEARNING FRAMEWORK

A network convolutional neural network model facing backend is proposed. Keras is an advanced deep neural network architecture with modularity, strong scalability and fast neural network deployment [14]. Figure 3 shows the configuration view of the WV-CNN model in Keras (image cited in A CNN-LSTM Architecture for Marine Vessel Track Association Using Automatic) Identification System (AIS) Data. Keras uses numpy, nltk, gensim and other processing software to implement three steps of credit information: text preprocessing, word vectorization, model training and credit evaluation. Keras is not only the execution architecture of high-level deep neural networks, but also for various types of models and functions, joint declaration functions, etc., organically combines the data obtained in the process of text preprocessing, word vectorization, training and evaluation, and finally builds a complete WV-CNN model [15]. In addition, model training and credit evaluation are integrated into Keras based on the configuration graph, text preprocessing, word vectorization, which not only simplifies the intermediate links of credit evaluation, but also can start from the internal understanding of credit evaluation, so as to better explore the independence of credit evaluation and lay the foundation for the subsequent research on credit evaluation.

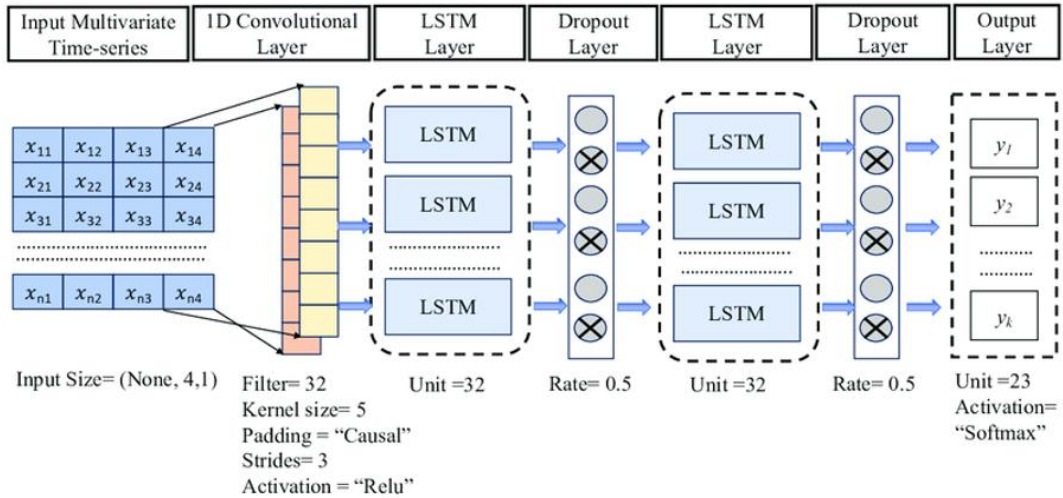

Fig. 3. Deployment diagram of WV-CNN model in Keras

## VI. Experiment

In order to test the correctness and superiority of the algorithm proposed in this paper, two general public databases, MSCOCO and Flickr30K, will be used to test the algorithm.

### A. Test use cases

The database of image object recognition and image scene description based on MSCOCO developed by Microsoft now has three versions, 2020, 2021 and 2022, including more than 80,000 training sets, 40,000 test sets and 40,000 images, covering a wide range of target types[16]. Each of the images collected contained five artificially labeled sentences. Most of the images in Coco database come from the real world, with more complex scenes, a large number of objects on the image, with an average of 7.7 per image, and more stringent evaluation indicators. Flickr, Yahoo's photo-sharing platform, offers more than 30,000 photos, mostly of people doing the same thing. Compared with MSCOCO, Flickr30K's biggest disadvantage is its small sample size, so this paper only uses it as a test sample. The dataset is processed using the Linked Data approach to unify diverse data formats, a key factor in academic research[17]. This structured methodology enables researchers to cross-reference information, enhancing interoperability between various datasets. This capability is especially advantageous in machine learning and artificial intelligence applications, where high-quality data is crucial for training models and obtaining accurate predictions.

### B. System hardware configuration and training parameter setting

The computer used in this article is a dedicated server for deep learning. Due to the use of a single GPU, the testing speed is lower, so during the testing process, this article will use the MSCOCO test set, which has a total of 883,242 months. In addition, the minimum batch size selected is 64, and the learning decay rate is 0.85.

### C. Test results and analysis

In this paper, MSCOCO and Flickr30k test sets are used to test the new mathematical models, and MSCOCO database is used to compare the scores of each model (Table 1 and Table 2).

TABLE I. COMPARISON OF SCORES OF DIFFERENT MODELS ON MSCOCO DATASET

| Model | B-1 | B-3 | B-4 | CIDEr |
|---|---|---|---|---|
| DeepVS | 0.658 | 0.378 | 0.282 | 0.850 |
| NIC | 0.722 | 0.407 | 0.317 | 0.941 |
| L R CN | 0.662 | 0.385 | 0.276 | - |
| Hard-Att | 0.714 | 0.386 | 0.277 | 0.862 |
| MSCap | 0.732 | 0.406 | 0.310 | 0.920 |
| E R D | 0.743 | 0.404 | 0.306 | 0.961 |
| OurModel | 0.740 | 0.426 | 0.332 | 0.943 |

TABLE II. COMPARISON OF SCORES OF DIFFERENT MODELS ON THE FLICKR30 DATASET

| Model | B-1 | B-3 | B-4 | CIDEr |
|---|---|---|---|---|
| DeepVS | 0.563 | 0.240 | 0.157 | 0.248 |
| NIC | 0.658 | 0.275 | 0.179 | - |
| ATT-FCN | 0.643 | 0.319 | 0.230 | - |
| OurModel | 0.650 | 0.318 | 0.235 | 0.353 |

As can be seen from the above table, compared with existing algorithms, the performance of this algorithm on MSCOCO and Flickr30k is significantly improved. In the training set of COCO, Tensorboard was used to observe the change of loss in the test over time, and the results were shown in Figure 4 .

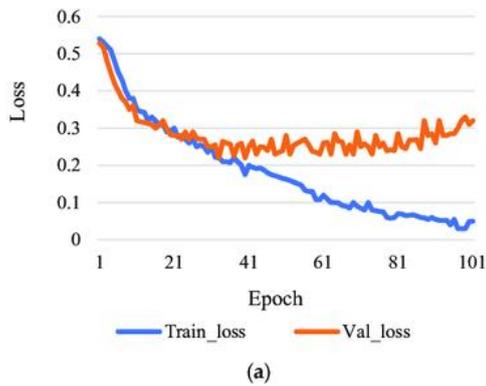
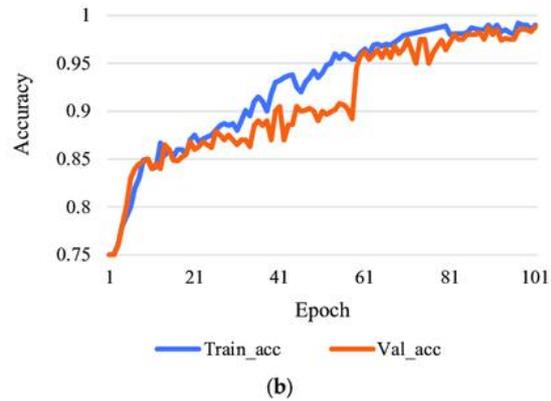

Fig. 4.   Changing trend of training loss degree

As can be seen from Figure 4, the LOSS value is still decreasing overall after experiencing some ups and downs. Experimental results show the effectiveness of this method.

## VII.  CONCLUSION

This research has delved into the optimization of image representation through the integration of attention mechanisms and multimodal data, culminating in a refined attribute model that encapsulates both semantic and latent image layers. By employing the Word2Vec methodology coupled with a word embedding convolutional neural network, the system efficaciously translates discrete feature sets into continuous attributes. This transformation represents a significant reduction in the complexity of feature preprocessing, while directly assessing the veracity of absent image features through advanced natural language processing techniques. In the pursuit of a truly autonomous image description generation, it is envisaged that further enhancements in the training process of the hierarchical LSTM networks, as well as the refinement of the attention scales, will play a pivotal role. The goal is to achieve an even closer mimicry of human-like descriptions, both in variety and complexity, ultimately contributing to the evolution of more intelligent and intuitive artificial intelligence systems. In summation, the novel approach developed in this project not only fortifies the bridge between visual perception and language processing but also illuminates the path toward eliminating the subjective bias in image feature evaluation, an advancement with promising implications for the domains of computer vision and machine learning.